\documentclass[conference]{IEEEtran}
\IEEEoverridecommandlockouts
\usepackage{cite}
\usepackage{amsmath,amssymb,amsfonts,amsthm}
\newtheorem{theorem}{Theorem}

\theoremstyle{definition} 
\newtheorem{definition}{Definition}
\usepackage{algorithmic}
\usepackage{tabu} 
\usepackage[linesnumbered,ruled,vlined]{algorithm2e}
\usepackage{graphicx}
\usepackage{subcaption}
\usepackage{textcomp}
\usepackage{xcolor}

\def\BibTeX{{\rm B\kern-.05em{\sc i\kern-.025em b}\kern-.08em
    T\kern-.1667em\lower.7ex\hbox{E}\kern-.125emX}}
\begin{document}

\title{UAV-Assisted Multi-Task Federated Learning with Task Knowledge Sharing\\
}

\author{\IEEEauthorblockN{Yubo Yang, Tao Yang, Xiaofeng Wu, and Bo Hu}
\IEEEauthorblockA{School of Information Science and Technology, Fudan University, Shanghai, China \\
Email: yangyb24@m.fudan.edu.cn, \{taoyang, xiaofengwu, bohu\}@fudan.edu.cn}
}

\maketitle

\begin{abstract}
The rapid development of Unmanned aerial vehicles (UAVs) technology has spawned a wide variety of applications, such as emergency communications, regional surveillance, and disaster relief. Due to their limited battery capacity and processing power, multiple UAVs are often required for complex tasks. In such cases, a control center is crucial for coordinating their activities, which fits well with the federated learning (FL) framework. However, conventional FL approaches often focus on a single task, ignoring the potential of training multiple related tasks simultaneously. In this paper, we propose a UAV-assisted multi-task federated learning scheme, in which data collected by multiple UAVs can be used to train multiple related tasks concurrently. The scheme facilitates the training process by sharing feature extractors across related tasks and introduces a task attention mechanism to balance task performance and encourage knowledge sharing. To provide an analytical description of training performance, the convergence analysis of the proposed scheme is performed. Additionally, the optimal bandwidth allocation for UAVs under limited bandwidth conditions is derived to minimize communication time. Meanwhile, a UAV-EV association strategy based on coalition formation game is proposed. Simulation results validate the effectiveness of the proposed scheme in enhancing multi-task performance and training speed.
\end{abstract}


\section{Introduction}
Unmanned Aerial Vehicle (UAV)-assisted networks are seen as an important enabler of post-5G and upcoming 6G networks. Due to their flexibility and line-of-sight (LOS) communication links, UAVs have great potential in providing emergency communications and wide-area on-demand data collection\cite{b0}. In addition, UAVs can be used to train popular machine learning models on-board, which is crucial for tasks such as trajectory planning and object recognition\cite{b1}. In a multi-UAV system, UAVs operate in a coordinated manner to support reliable and efficient communications or complete specific tasks\cite{b2}\cite{b3}. UAV coordination is typically assisted by a ground base station. For instance, in an emergency rescue scenario with poor network coverage, a swarm of UAVs is controlled by a ground emergency vehicle (EV). However, since UAVs are deployed in various areas and often perform data collection dynamically\cite{b4}, given the limited communication resources and energy capacity of UAVs, it is unrealistic to collect and transmit raw data from each UAV to the EV, as well as to frequently shuttle between the incident scene and the EV for charging. Given these challenges, federated learning (FL) has emerged as a promising framework for distributed data processing and model training. Taking advantage of the capabilities of the UAV data platform, it enables collaborative learning among multiple UAVs, thereby enhancing model training efficiency and accuracy\cite{b5}.

Currently, most existing studies focus on single-task learning, while some multi-task FL either treats each individual model as a separate task \cite{b6}\cite{b7} or learns multiple unrelated tasks concurrently \cite{b8}\cite{b9}. This differs from the scenario considered in this paper. Actually, the data collected by UAVs may be used for multiple related tasks. For example, in emergency rescue scenarios, UAV images can support tasks like disaster level identification, crowd density estimation, and road feasibility analysis. Therefore, the multi-task scheme considered in this paper refers to using data collected by UAVs to complete multiple related but different tasks simultaneously. Using the traditional single-task learning framework to learn each task independently is inefficient\cite{b10}. Since multiple tasks are trained on the same image dataset, they share common underlying features and may have potential correlations, which allows for effective knowledge sharing among tasks and enables a reduction in resource consumption for training.

Studies have shown that sharing feature extractors across correlated tasks can enhance both the training speed and robustness of models\cite{b11}. However, these findings are limited to centralized training scenarios. In our scenario, each UAV is deployed in a different area or assigned a distinct but related task, leading to variations in local data collection. Integrating data features collected by multiple UAVs assigned to different tasks significantly enhances the robustness and generalization of the feature extractors. This approach allows the extractor to learn from a broader range of areas, thus improving feature extraction performance, accelerating model convergence, and boosting overall performance through mutual reinforcement among tasks\cite{b12}. In summary, the contributions of this paper are as follows:
\begin{itemize}
    \item We propose a novel UAV-assisted multi-task federated learning scheme, where the data collected by the UAVs can be used to train multiple related tasks, and the system performance and robustness can be improved by sharing knowledge among related tasks.
    \item To balance the task performance and enhance the overall effectiveness of multiple tasks, we propose a task attention mechanism to capture the dynamic importance of each task. Specifically, we introduce the task shapley value to facilitate knowledge sharing among tasks.
    \item Considering the dynamic environment of the network and resource constraints, we derived an expression for optimal bandwidth allocation for UAVs and introduced a UAV-EV association mechanism based on coalition formation game. Simulation results demonstrate the effectiveness of the proposed scheme.
\end{itemize}
\section{System Model}
\begin{figure}[h]
    \centering
    \includegraphics[width=0.45\textwidth]{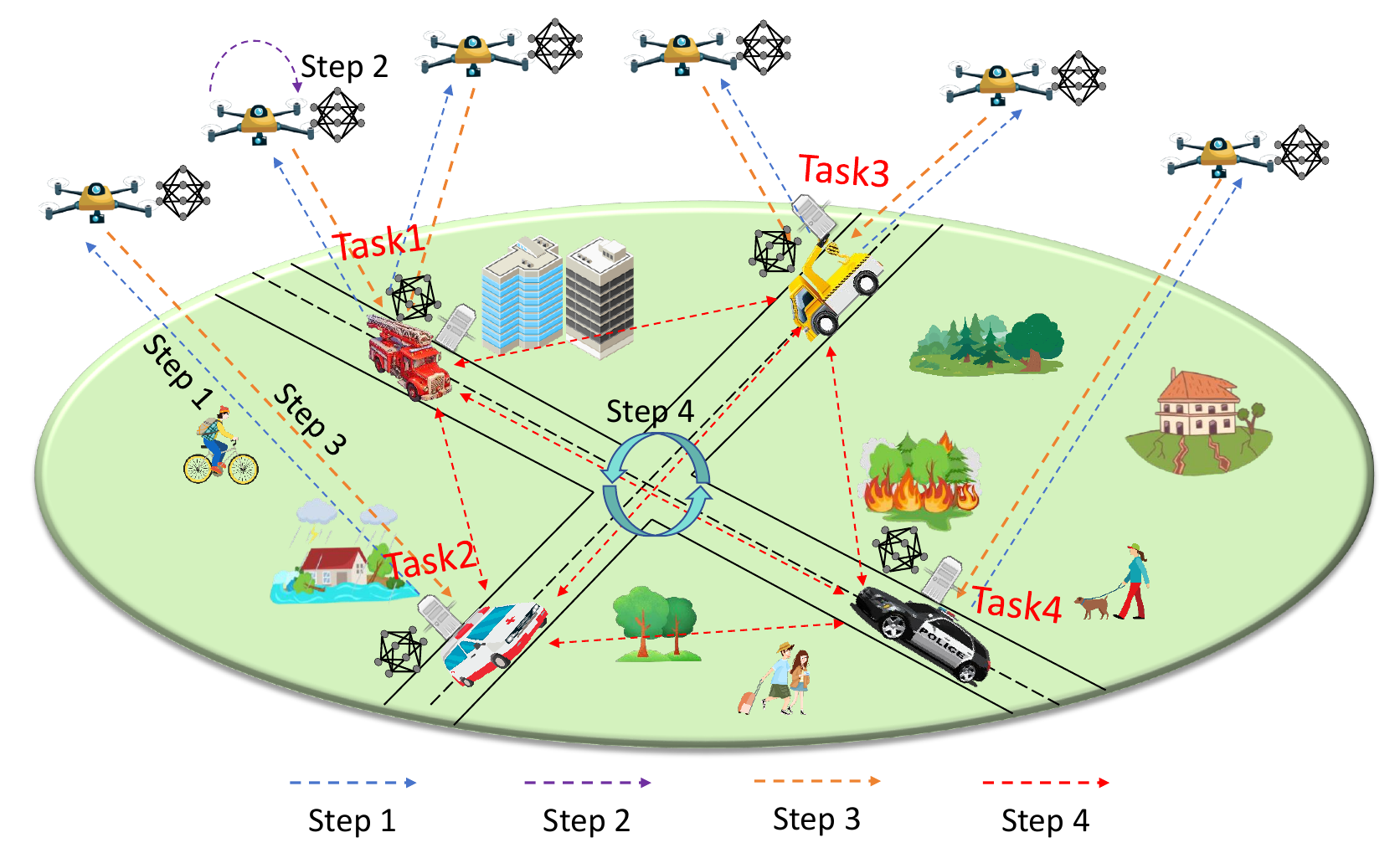}  
    \caption{UAV assisted multi-task federated learning system.}
    \label{fig:shiyitu}
\end{figure}

\subsection{Federated Learning System}
The system consists of $M$ EVs indexed by $\mathcal{M}=\{1,2, \cdots, M\}$ and $N$ UAVs indexed by $\mathcal{N}=\{1,2, \cdots, N\}$. Each EV is assigned a different task and can interact with other EVs to share task knowledge. Additionally, each EV can collect a small amount of data for model validation. Each UAV $n$ has a local dataset $\mathcal{D}_n$ with $D_n=\left|\mathcal{D}_n\right|$ data samples, and the data can be used for $M$ tasks. Since the same image dataset is used for multiple tasks, the underlying image features required by multiple tasks may be correlated. To enhance feature extraction performance, we allow multiple tasks to share feature extractors while maintaining task-specific predictors. The entire dataset is
denoted by $\mathcal{D}=\cup\left\{\mathcal{D}_n\right\}_{n=1}^N$ with a total number of samples $D=\sum_{n=1}^N D_n$. Given a data sample $(\boldsymbol{x}, \boldsymbol{y}) \in \mathcal{D}$, where $\boldsymbol{x} \in \mathbb{R}^d$
is the $d$-dimensional input data vector, $\boldsymbol{y}=[y_1,y_2,\cdots,y_M] \in \mathbb{R}^M$ is the corresponding $M$-dimensional ground-truth label and each dimension corresponds to a specific task. Let $f_{m,n}(\boldsymbol{x},\boldsymbol{y}; \boldsymbol{w}_{m,n})$ denote the sample-wise loss function for task $m$ on UAV $n$, $\boldsymbol{w}_{m,n}$ can be split into two parts $[\boldsymbol{w}_{m,n}^s,\boldsymbol{w}_{m,n}^u]$, where $\boldsymbol{w}_{m,n}^s$ represents the shared feature extractor and $\boldsymbol{w}_{m,n}^u$ represents the unique layer of task $m$. Thus, the local loss function that UAV $n$ measures the model error for task $m$ on its local dataset is given by
\begin{equation}
     {F_{m,n}(\boldsymbol{w}_{m,n})=\frac{1}{D_{n}}\sum_{(\boldsymbol{x},\boldsymbol{y})\in\mathcal{D}_{n}}f_{m,n}(\boldsymbol{x},\boldsymbol{y};\boldsymbol{w}_{m,n})}
\end{equation}
Accordingly, the global loss function for task $m$ is
\begin{equation}
F_m(\boldsymbol{w}_m)=\sum\nolimits_{n=1}^N \frac{D_n}{D}  F_{m,n}(\boldsymbol{w}_{m})
\end{equation}

We use $\boldsymbol{w}=[\boldsymbol{w}_1,\boldsymbol{w}_2,\cdots\boldsymbol{w}_M]$ to represent the set of all task parameters. For simplicity, we omit the subscript \(m\) from \(f\) and use \(f_{n}(\boldsymbol{x}, \boldsymbol{y}; \boldsymbol{w}_{m,n})\) to denote the loss function of task \(m\) on UAV \(n\). Each UAV can participate in different tasks in different rounds. However, due to the limited computing resources and energy, each UAV can only participate in one training task per round. Considering the goal to improve the average performance of all tasks, the objective function of the entire network is
\begin{equation}
    F(\boldsymbol{w})=\frac{1}{M} \sum\nolimits_{m=1}^M  F_{m}(\boldsymbol{w}_{m})
\end{equation}

To achieve the objective outlined in (3), the training steps of the proposed scheme are given as follows:

$\bullet$ \textbf{(Step 1 UAV-EV association and Task Model Broadcast)} At the beginning of each round, each UAV is assigned to an EV through the UAV-EV association algorithm. Each EV broadcasts its latest model parameters $\boldsymbol{w}_{m,t}$ to its associated UAVs. Let $\beta_{m,n, t} \in\{0,1\}$ denote the association indicator of UAV $n$ in rount $t$, where $\beta_{m,n, t} =1$ indicates that
UAV $n$ is associated with EV $m$,  $\beta_{m,n, t} =0$ otherwise.  We use $\boldsymbol{\beta_{m,t}} =[\beta_{m,1,t},\beta_{m,2,t},\cdots,\beta_{m,N,t}]^T\in\mathbb{R}^N , \forall m\in\mathcal{M}$  to represent the UAV association of task $m$ in round $t$. And $\boldsymbol{\beta}_t=[\boldsymbol{\beta_{1, t}}; \boldsymbol{\beta_{2, t}}; \cdots; \boldsymbol{\beta_{M, t}}]^T$ can be used to represent the overall UAV-EV association in round $t$.

$\bullet$ \textbf{(Step 2 Local Model Training)} Each UAV $n$ initializes $\boldsymbol{w}_{m,n,t}^0=\boldsymbol{w}_{m,t}$ and performs $K$ steps stochastic gradient descent (SGD) to update the shared layer and the task-specific layer.
$\boldsymbol{w}_{m,n,t}^{s}$ and $\boldsymbol{w}_{m,n,t}^{u}$ are updated as follows:
\begin{equation}
   \boldsymbol{w}_{m,n,t}^{s,k+1}=\boldsymbol{w}_{m,n,t}^{s,k}-\eta\nabla_{s}\widetilde{f}_{n}(\boldsymbol{w}_{m,n,t}^{s,k},\boldsymbol{w}_{m,n,t}^{u,k})
\end{equation}
\begin{equation}
   \boldsymbol{w}_{m,n,t}^{u,k+1}=\boldsymbol{w}_{m,n,t}^{u,k}-\eta\nabla_{u}\widetilde{f}_{n}(\boldsymbol{w}_{m,n,t}^{s,k},\boldsymbol{w}_{m,n,t}^{u,k})
\end{equation}
where $\boldsymbol{w}_{m,n,t}^{s,k}$ and $\boldsymbol{w}_{m,n,t}^{u,k}$ are the feature extractor layer and task-specific layer parameters of task $m$ on UAV $n$ in the $k$-th local iteration of the $t$-th round, respectively. $\eta>0$ is the learning rate.

In (4), the stochastic gradient $\nabla_{s}\widetilde{f}_{n}(\boldsymbol{w}_{m,n,t}^{s,k},\boldsymbol{w}_{m,n,t}^{u,k})$ is given by
$\nabla_{s}\widetilde{f}_{n}(\boldsymbol{w}_{m,n,t}^{s,k},\boldsymbol{w}_{m,n,t}^{u,k})=\frac{1}{L_b} \sum_{(\boldsymbol{x},\boldsymbol{y}) \in \zeta_{n,t}^h} \nabla_s f_{n}\left(\boldsymbol{x},\boldsymbol{y} ; \boldsymbol{w}_{m,n,t}^{k}\right)$,
where $\zeta_{n,t}^k$ is a mini-batch data uniformly sampled from $D_n$
with $L_b=\left|\zeta_{n,t}^k\right|$ data samples. The update form of $\nabla_{u}\widetilde{f}_{n}(\boldsymbol{w}_{m,n,t}^{s,k},\boldsymbol{w}_{m,n,t}^{u,k})$
is similar and is omitted here.

$\bullet$ \textbf{(Step 3 Model Update $\&$ Parameter Uploading )} After local model training, each UAV $n$ uploads its cumulative local stochastic gradient $\tilde{\boldsymbol{G}}_{m,n,t}$ to its corresponding EV.
$\tilde{\boldsymbol{G}}_{m,n,t}$ is given by
$\tilde{\boldsymbol{G}}_{m,n,t}=\sum_{k=0}^{K-1} \nabla\widetilde{f}_{n}(\boldsymbol{w}_{m,n,t}^{s,k},\boldsymbol{w}_{m,n,t}^{u,k})=\frac{1}{\eta}\left(\boldsymbol{w}_{m,n,t}^0-\boldsymbol{w}_{m,n,t}^{K}\right)$
Note that the cumulative gradient can be expressed as $\tilde{\boldsymbol{G}}_{m,n,t}=[\tilde{\boldsymbol{G}}_{m,n,t}^{s},\tilde{\boldsymbol{G}}_{m,n,t}^{u}]$, where $\tilde{\boldsymbol{G}}_{m,n,t}^{s}$ and $\tilde{\boldsymbol{G}}_{m,n,t}^{u}$ represent the feature extractor gradient and task-specific layer gradient respectively.

$\bullet$ \textbf{(Step 4 EV Model Aggregation and Knowledge Sharing)}
Each EV aggregates the latest gradient and updates the global model of task $m$. Specifically, the feature extractor is updated as
$\hat{\boldsymbol{w}}_{m,t+1}^{s}=\boldsymbol{w}_{m,t}^{s}-\eta\frac{ {\textstyle \sum\nolimits _{n=1}^{N}}\beta_{m,n,t}D_n\tilde{\boldsymbol{G}}_{m,n,t}^{s} }{\sum\nolimits_{n=1}^{N}\beta_{m,n,t}D_n}$
and the task predictor is updated as
$\boldsymbol{w}_{m,t+1}^{u}=\boldsymbol{w}_{m,t}^{u}-\eta\frac{ {\textstyle \sum\nolimits _{n=1}^{N}}\beta_{m,n,t}D_n\tilde{\boldsymbol{G}}_{m,n,t}^{u} }{\sum\nolimits_{n=1}^{N}\beta_{m,n,t}D_n}$. Subsequently, the EVs interact to share knowledge by transmitting $\hat{\boldsymbol{w}}_{m,t+1}^{s}$ to one another for further fusion of the feature extractor parameters. Given that EVs possess relatively sufficient transmission resources and computing capabilities, we disregard any delays and energy consumption associated with this parameter exchange. The fused feature extractor can then be expressed as: $\boldsymbol{w}_{t+1}^{s}=\frac{{\textstyle \sum\nolimits _{m=1}^{M}}D_{m,t}\hat{\boldsymbol{w}}_{m,t+1}^{s} }{\sum\nolimits_{m=1}^{M}D_{m,t}}$, where $D_{m,t}=\sum\nolimits_{n=1}^{N} \beta_{m,n,t}D_n$.
\subsection{Computation Model}
Let $C_{m,n}$ denotes the number of CPU cycles required to process one data sample when UAV $n$ participates in the training of task $m$, $f_n$ represents the computation capability of UAV $n$. Thus, if UAV $n$ participate in the training of task $m$, the computational time of local training of is given by
$T_{m,n, t}^{comp}=\frac{K L_b C_{m,n}}{f_n}$ and the corresponding energy consumption of UAV $n$ is $E_{m,n, t}^{comp}=\varsigma KL_b C_{m,n}\left(f_n\right)^2$, where $\varsigma$ is the energy coefficient, which depends on the chip architecture. 
\subsection{Communication Model}
The air-to-ground communication link is modeled as a probabilistic superposition of LOS and NLOS channels based on the method in \cite{b13}. For the sake of brevity, the details are omitted here. 
We consider using the frequency division multiple access (FDMA) techniques for communication. Each UAV can be allocated a certain proportion $\gamma _{n, t}$ of uplink bandwidth resources. Denote $p_{n,t}$
as the transmit power of UAV $n$ in round $t$, and it has a maximum transmission power limit $p_{n,max}$. Thus, the achievable transmission rate between UAV $n$ and EV $m$ in round $t$ is
$r_{m,n, t}=\gamma_{n, t} B \log _2\left(1+\frac{p_{n, t} h_{m,n, t}}{\gamma_{n,t}B N_0}\right)$. $N_0$ is the noise power spectral density. Let $Q_m$ denote the size of the gradient for task $m$, if UAV $n$ is scheduled to participate in task $m$ in round $t$, its transmission time is given by $T_{m,n, t}^{comm}=\frac{Q_m}{r_{m,n, t}}$
The corresponding energy consumption of UAV $n$ for transmission is
$E_{n, t}^{comm}=p_{n, t} T_{m,n, t}^{comm}$.

\section{PROBLEM FORMULATION AND CONVERGENCE ANALYSIS}
\subsection{Problem Formulation}
In order to minimize the expected average global loss of all tasks and reduce the total training time subject to the constrained resources of UAVs, we need to jointly optimize UAV-EV association, bandwidth allocation, and UAV transmission power. The problem is formulated as follows:
\begin{subequations}
\renewcommand{\theequation}{19\alph{equation}}
\begin{align}
P_1: &\min_{\left\{\boldsymbol{\beta}_t,\boldsymbol{\gamma}_t,\boldsymbol{p}_t\right\}_{t=0}^{T-1}}    \lambda \mathbb{E}[F(\boldsymbol{w}_T)] + (1-\lambda) T_{\text{total}} \tag{6}\\
\text{s.t. } & E_{n,t}^{comp} + E_{n,t}^{comm} \leq E_{n,\text{max}}, \quad \forall n \in \mathcal{N}, \forall t, & \tag{6a} \label{eq:total_energy} \\
& \beta_{m,n,t} \in \{0,1\}, \quad \forall m \in \mathcal{M}, \forall n \in \mathcal{N}, \forall t, & \tag{6b} \label{eq:beta_binary} \\
& \sum\nolimits_{m=1}^M \beta_{m,n,t} = 1, \quad \forall n \in \mathcal{N}, \forall t, & \tag{6c} \label{eq:beta_sum} \\
& \sum\nolimits_{n=1}^N \beta_{m,n,t} \geq \delta_m, \quad \forall m \in \mathcal{M}, \forall t, & \tag{6d} \label{eq:delta_min} \\
& 0\le\gamma_{n,t}\leq 1, \quad \forall n \in \mathcal{N}, \forall t, & \tag{6e} \label{eq:delea_min}  \\
& \sum\nolimits_{n=1}^{N} \gamma_{n,t}\le 1, \quad \forall t, & \tag{6f} \label{eq:sum_gamma} \\
& 0 \leq p_{n,t} \leq p_{n,\text{max}}, \quad \forall n \in \mathcal{N}, \forall t. & \tag{6g} \label{eq:power_bounds}
\end{align}
\end{subequations}

The constraint in (6a) ensures that the energy consumption of each UAV in each round cannot exceed the maximum energy limit $E_{n,\text{max}}$ per round. Constraints (6b)-(6d) imposes restrictions on UAV-EV associations. Constraints (6e) and (6f) regulate the bandwidth allocation, and (6g) limits the transmission power of each UAV.

Since the objective function of problem \(P_1\) involves the neural network's loss function \(F(\boldsymbol{w}_t)\), the factors affecting the network's performance are still unknown. Consequently, problem \(P_1\) is not yet solvable. To address this, we first conduct the convergence analysis of the proposed scheme to identify the factors affecting task performance, and then formulate the problem into a solvable suboptimal form.

\subsection{Convergence Analysis}
\noindent$\textbf{\textit{Assumption 1}}$. For each UAV $n$ participating in each task $m$, the stochastic gradient is unbiased and variance bounded
\begin{equation}
   \mathbb{E} \left[\nabla_{s}\widetilde{f}_{n}(\boldsymbol{w}_{m,n,t}^{k})\right ]=\nabla_{s}f_{n}(\boldsymbol{w}_{m,n,t}^{k})
\end{equation}
\begin{equation}
\mathbb{E} \left[\left \| \nabla_{s}\widetilde{f}_{n}(\boldsymbol{w}_{m,n,t}^{k})-\nabla_{s}f_{n}(\boldsymbol{w}_{m,n,t}^{k})\right \|^2\right ]\leq \sigma_s^2  
\end{equation}
The task-specific layer properties are similar, which are omitted here.

\noindent$\textbf{\textit{Assumption 2}}$. The expected squared norm of local gradients for each task $m$ and UAV $n$ is uniformly bounded
\begin{equation}
\mathbb{E}\left[\left\|\nabla_{s}f_{n}(\boldsymbol{w}_{m,n,t}^{k})\right\|^2\right]\leq \epsilon_s  ^2,
\text{ } \mathbb{E}\left[\left\|\nabla_{u}f_{n}(\boldsymbol{w}_{m,n,t}^{k})\right\|^2\right]\leq \epsilon_u ^2.
\end{equation}
\noindent$\textbf{\textit{Assumption 3}}$. For each task $m$ and UAV $n$, there exist $L_s,L_u,L_{su}$ and $L_{us}$ such that:

\noindent$\bullet$$\text{  }\nabla_s f_{n}(\boldsymbol{w}_{m,t}^{s},\boldsymbol{w}_{m,t}^{u})$ is $L_s$-Lipschitz continuous with $\boldsymbol{w}_{m,t}^{s}$ and $L_{su}$-Lipschitz continuous with $\boldsymbol{w}_{m,t}^{u}$:
\begin{equation}
    ||\nabla_s f_{n}(\bar{\boldsymbol{w}}^{s}, \boldsymbol{w}_{m,t}^{u})-\nabla_s f_{n}(\check{\boldsymbol{w}}^{s},\boldsymbol{w}_{m,t}^{u})||\le L_s||\bar{\boldsymbol{w}}^{s}-\check{\boldsymbol{w}}^{s}||
\end{equation}
\begin{equation}
    ||\nabla_s f_{n}(\boldsymbol{w}_{m,t}^{s}, \bar{\boldsymbol{w}}^{u})-\nabla_s f_{n}(\boldsymbol{w}_{m,t}^{s},\check{\boldsymbol{w}}^{u})||\le L_{su}||\bar{\boldsymbol{w}}^{u}-\check{\boldsymbol{w}}^{u}||
\end{equation}

\noindent$\bullet$$\text{  }\nabla_u f_{n}(\boldsymbol{w}_{m,t}^{s},\boldsymbol{w}_{m,t}^{u})$ is $L_u$-Lipschitz continuous with $\boldsymbol{w}_{m,t}^{u}$ and $L_{us}$-Lipschitz continuous with $\boldsymbol{w}_{m,t}^{s}$, the detail is omitted here.

Based on the above assumptions, we give the upper bound of the difference in loss function between two consecutive rounds of task $m$:
\begin{theorem}
Given the UAV-EV association \(\beta_t\) in round \(t\), when the learning rate \(\eta < \frac{2}{K^2(L_u + 2L_{su})}\), the difference in the loss function for task \(m\) between two consecutive rounds is bounded by:
\begin{equation}
\begin{aligned}
&F_{m}(\boldsymbol{w}_{t+1}^{s},\boldsymbol{w}_{m,t+1}^{u})-F_{m}(\boldsymbol{w}_{t}^{s},\boldsymbol{w}_{m,t}^{u})\\
&\leq -\frac{\eta K}{2} \left [  \mathbb{E}\left \| \nabla_{s}F_{m}(\boldsymbol{w}_{m,t})\right \|^2+ \mathbb{E}\left \| \nabla_{u}F_{m}(\boldsymbol{w}_{m,t})\right \|^2\right ]\\
&+\frac{\left (D-\sum\nolimits_{n=1}^{N}\beta_{m,n,t}D_n\right )^2\Omega_1}{D^2}\\
&+\frac{\left (D-\sum\nolimits_{n=1}^{N}\beta_{m,n,t}D_n\right )\Omega_2}{D}+\Omega_3
\end{aligned}
\end{equation}
where $ \Omega_1=(4\eta K+16K^2L_{su})\epsilon_s^2+4\eta K\epsilon_u^2$, $\Omega_2=8K^2\sigma_sL_{su}\epsilon _s+4K\epsilon _s^2$,  $\Omega_3=K\sigma _s(\epsilon _s+2K\sigma _sL_{su})+\frac{\eta K}{2}\Omega_4$ and $
\Omega_4= \left ( \frac{\left (K+1\right )\left (L_u \epsilon_u+L_{us} \epsilon_{s}\right )}{2}+\sigma _u \right )^2 +\left ( \frac{\left (K+1\right )\left (L_s \epsilon_s+L_{su} \epsilon_{u}\right )}{2}+\sigma _s \right )^2 
$
\end{theorem}
The proof is omitted here due to space limitations.

According to Theorem 1, it can be seen that task learning performance is mainly affected by the amount of associated data. Due to the dynamic UAV-EV association and the varying complexity of tasks, the training progress differs from task to task. Given the diminishing marginal effect observed in neural network training, improving overall task performance calls for greater emphasis on tasks with slower training progress. Besides, due to the differences in the quality of UAV data collected in different regions, the feature extractors of EVs associated with these high-quality data UAVs usually have better feature extraction capabilities. Through feature extractor fusion, these EVs usually bring higher performance improvements to other EVs. Therefore, EVs with a strong performance history should receive greater focus. Based on the above analysis, each task should have a dynamic task weight in different rounds. Therefore, to address problem P1, we reformulate it into the following form.
\begin{subequations}
\renewcommand{\theequation}{13\alph{equation}}
\begin{align}
P_2: &\max_{\boldsymbol{\beta}_t,\boldsymbol{\gamma}_t,\boldsymbol{p}_t}   \lambda \sum_{m=1}^{M} \alpha_m\sum_{n=1}^{N}\beta _{m,n,t}\rho_n -(1-\lambda)T_t \tag{13}\\
\text{s.t. } &(6a-6g)
\end{align}
\end{subequations}
where $\rho_n=\frac{D_n}{D}$ represents the sample size ratio of UAV $n$ and this problem will be addressed in section IV.
\section{Proposed Algorithm}
\subsection{Task Attention Mechanism}
In this section, we propose a task-level attention mechanism to dynamically characterize the weights of each task in each round.
\subsubsection{Task Performance Balance}
In a multi-task learning system, achieving a balanced performance across tasks is key to improving the system's overall performance. Excessive performance differences among tasks may lead to longer delays and unnecessary waste of resources. To this end, we propose a dynamic weight adjustment mechanism based on historical loss, Specifically, each EV $m$ maintains the parameter $\Gamma_{m,t}$ to record the current weighted cumulative loss value of its own task and
$\Gamma_{m,t}=\varpi\Gamma_{m,t-1}+(1-\varpi)loss_{m,t-1}$,
where $loss_ {m, t-1}$ represents the loss of task $m$ when the $t-1$ round of training is completed, and $\varpi$ is the parameter that balances the importance of historical cumulative loss and current loss.

Then, the loss based weight $\tilde{\Psi}_{m,t}$ of each task $m$ can be obtained as
$\tilde{\Psi}_{m,t}=\frac{e^{\Gamma_{m,t}}}{\sum_{m=1}^Me^{\Gamma_{m,t}}}$.

\subsubsection{Task Shapley Value}
In this part, the concept of the Shapley value is used to quantify the marginal contribution of task \(m\) in enhancing the performance of all tasks. We use $C$ to represent the EV subset, and utility function $v_m(C)$ represents the accuracy of task $m$ after updating the feature extractor with the gradient of EV contained in $C$. Then, the shapley value of task $m$ in round $t$ can be calculated as
\begin{equation}
\phi_{m,t}=\frac{1}{M^2}\sum_{\substack{i=1}}^{M}\sum_{j=0}^{M-1}\sum_{\substack{|C|=j,\\m\notin C}}\frac1{\binom{M-1}j}(v_i(C\cup\{m\})-v_i(C)) 
\end{equation}
Then, the cumulative Shapley value of each task $m$ is updated to
$\mathcal{I}_{m,t}=\kappa \mathcal{I}_{m,t-1}+(1-\kappa  )\phi_{m,t}$, $\kappa$ is a balancing factor similar to $\varpi$.
To promote knowledge sharing among tasks and accelerate the training process of each task, tasks with high shapley values should receive greater focus, so the task shapley value based weight $\hat{\Psi}_{m,t}$ can be calculated as
$\hat{\Psi}_{m,t}=\frac{e^{\mathcal{I}_{m,t}}}{\sum_{m=1}^Me^{\mathcal{I}_{m,t}}}$.

Finally, we can get the weight of each task in round $t$ based on the task attention mechanism as
$\Psi_{m,t}=\frac{\tilde{\Psi}_{m,t}+\hat{\Psi}_{m,t}}{\sum_{m=1}^M(\tilde{\Psi}_{m,t}+\hat{\Psi}_{m,t})}$.
\subsection{Optimal Bandwidth Allocation}
Given the UAV-EV association $\boldsymbol{\beta_{t}}$, problem $P_2$ can be simplified as
\begin{subequations}
\renewcommand{\theequation}{15\alph{equation}}
\begin{align}
&\min_{\gamma_{t}, p_{t}} T_t \tag{15}\\
\text{s.t. } 
& (6a),(6e-6g)\tag{15a}
\end{align}
\end{subequations}

We first determine the transmission power $p_{n,t}$ for each UAV. From the previous communication model, we know that \( \frac{\partial r_{n,t}}{\partial p_{n,t}} > 0 \). To reduce transmission time, each UAV should operate at its highest possible transmission rate within the limits of its energy constraints. Therefore, the transmission power for each UAV is set to \( p_{n,t} = p_{n,\text{max}} \). Consequently, problem (14) can be rewritten as:
\begin{subequations}
\renewcommand{\theequation}{16\alph{equation}}
\begin{align}
&\min_{\gamma_{t}} T_{t}^{max} \tag{16}\\
\text{s.t. } 
& T_{n,t}^{comp}+T_{n,t}^{comm}\leq T_t^{max},\forall n\in\mathcal{N},\forall t  & \tag{16a} \label{eq:gamma range1} \\
& (6e-6f) & \tag{16b}
\end{align}
\end{subequations}
where $T_t^{\text{max}} = \max_{n} \{T_{n,t}^{\text{comp}} + T_{n,t}^{\text{comm}}\}
$ and the optimal solution of (16) is presented in Theorem 2 as follows. 
\begin{theorem}
Given the UAV-EV association $\beta_t$, the optimal bandwidth allocation ratio for each UAV is
\begin{equation}
    \gamma _{n,t}=\frac{d_{n,t}\ln2}{\left(T_t^*-T_{n,t}^{\mathrm{comp}}\right)\left(\mathcal{W}\left(-\chi_{n,t}e^{-\chi_{n,t}}\right)+\chi_{n,t}\right)}
\end{equation}
where $\chi_{n,t}\triangleq\frac{d_{n,t}N_0\mathrm{ln}2}{\left(T_t^*-T_{n,t}^\mathrm{comp}\right)P_{n,t}h_{n,t}^2}$, $\mathcal{W(\cdot)}$ is the Lambert-W function and $T_t^*$ is the optimal solution to problem $(16) $ and satisfies
\begin{equation}
\sum_{n=1}^N\frac{d_{n,t}\ln2}{\left(T_t^*-T_{n,t}^{\mathrm{comp}}\right)\left(\mathcal{W}\left(-\chi_{n,t}e^{-\chi_{n,t}}\right)+\chi_{n,t}\right)}=1
\end{equation}
\end{theorem}
The proof is omitted here due to space limitations.

Since the optimal bandwidth allocation requires solving the Lambert-$W$ function, to obtain a closed-form expression just through parameter iteration computation explicitly is almost impossible. Therefore, we adopt a bisection method to give the solution numerically.

\subsection{Optimal UAV-EV Association}
Once the association between the UAV and the EV is determined, the optimal bandwidth allocation ratio and the minimum round time $T_t^*$ are determined. Substituting the minimum time $T_t^*$ obtained in the previous section into equation (16), we can transform the problem as
\begin{subequations}
\renewcommand{\theequation}{19\alph{equation}}
\begin{align}
&\max_{\beta_{t}} \lambda \sum_{m=1}^{M} \alpha_m^t\sum_{n=1}^{N}\beta _{m,n}^t\rho_n -(1-\lambda)T^* \tag{19}\\
\text{s.t. } 
& (6b-6d) 
\end{align}
\end{subequations}

Since the objective function does not have a clear analytical expression at this time and the constraints contain binary constraint variables, the problem is a non-convex optimization problem. In order to optimize the association between UAVs and EVs in an efficient and feasible way, we use the coalition formation game to obtain a suboptimal solution to the problem.

First we introduce some essential concepts and definitions in the following part.
\begin{definition}
All $N$ UAVs can be divided into $M$ clusters, expressed as $\mathcal{S}=\left \{ S_i\mid i= 1,2,\cdots M\right \} $ and thus constitute a UAV-EV association strategy. We have $S_i \cap S_j=\phi, \forall i,j=1,2,\cdots M$ and $\bigcup_{i=1,\cdots M } S_i=\mathcal{N} $.
\end{definition}

\begin{definition}
Given the current UAV-EV association $\mathcal{S}$, the utility is given by (13) and denoted as $U(\mathcal{S})$.
\end{definition}

\begin{definition}
Given two different UAV-EV association strategies $S_1$ and $S_2$, we define a preference order as $S_1 \succ S_2$ if and only if $U(S_1)>U(S_2)$.
\end{definition}
\begin{definition}
A UAV transferring adjustment by $n$ means that UAV $n\in S_m$ with $\left | S_m \right | > \delta_m$ retreats its current training group $S_m$ and joins another training group $S_{-m}$. Causing the network utility to change from $U(S_1)$ to $U(S_2)$ and $U(S_2)>U(S_1)$.
\end{definition}
\begin{definition}
A UAV exchanging adjustment means that UAV $n_1\in S_{m_1}$ and $n_2\in S_{m_2}$ are switched to each other’s training group. Causing the network utility to change from $U(S_1)$ to $U(S_2)$ and $U(S_2)>U(S_1)$.
\end{definition}
\begin{definition}
A UAV-EV association strategy $S^*$ is at a stable system point if no EV $m$ and UAV $n$ can change its association strategy to obtain a higher global network utility.
\end{definition}

Based on the above definition, we can solve (19) by continuously adjusting the UAV-EV association policy $S$ to obtain a higher utility according to the preference order. The association policy adjustment will terminate when a stable association $S^*$ is reached, where no UAV in the system has the motivation to leave its currently associated EV and join another EV. 

 
After the UAV-EV association algorithm converges, the optimal UAV-EV association and resource allocation strategy will be broadcast to all UAVs. All the involved UAVs will participate in the task of its associated EV and execute local training. 
\section{Numerical Results}
We consider a cellular network with a coverage radius of 500$m$, where $N$ UAVs and $M$ EVs are randomly distributed in the network.
\subsection{Datasets and local models}
This work utilizes the MNIST dataset for simulation validation and some modifications are performed to make it suitable for multi-task training. The digits are first rotated, and the rotation angle is selected from the preset angle vector $[0,36,\cdots,324]$. In addition, the digits and background are dyed differently to simulate the different backgrounds of the images collected by UAVs. For the modified MNIST dataset, we use it to complete two tasks: digit recognition and rotation angle recognition. Both tasks require information about the shape and contours of the digits, so there are similarities in the underlying tasks. For convenience, we use the same predictor for both tasks, but the results in this article can be easily extended to the case of different predictors. The model shared layers contain 3 convolutional layers, each followed by a batch normalization layer. Each task retains its own 4 fully connected layers.
\subsection{Hyperparameters and baselines}
We set $N=10$, $M=2$, and distribute training data to each UAV based on two dirichlet distributions. The parameter $\alpha_1$ controls the size distribution of the UAV dataset, and the parameter $\alpha_2$ controls the proportional distribution of the digit categories and rotation angles. The remaining network parameters are shown in Table $\mathrm{I}$. The three baseline strategies are set as follows: \textbf{Strategy 1} shares the feature extractor, but UAV-EV associations are randomly associated. \textbf{Strategy 2} shares the feature extractor, but each UAV is associated with its nearest EV. \textbf{Strategy 3} does not share the feature extractor, and UAV-EV associations are randomly associated, which is similar to the FedAvg algorithm.
\addtolength{\topmargin}{0.03in}
\begin{table}
\begin{center}
\caption{Parameter settings}
\begin{tabu} to 0.45\textwidth {X[c]|X[b]|X[c]|X[m]}
\hline
Parameter & Value & Parameter & Value \\
\hline
$N$ & 10 & $M$ & 2 \\
$B$ & 2MHz &$L_b$& 500 \\
$N_0$ & -174dBm & $H$ & 100m \\
$\eta$ & 0.01 & $\kappa$ &0.8\\
$\varpi$ & 0.8& $f_n$ & $[1,2]$GHz \\
\hline
\end{tabu}
\end{center}
\end{table}

\subsection{Performance Comparison}
\textbf{Accuracy Comparison} 
In this part, energy and time constraints are not considered, so we set $\lambda=1$. Figure $2(a)$ shows the average accuracy of the two tasks as the training progresses over multiple rounds. The results show that the proposed algorithm is better than the three baseline strategies. When compared to Strategy 1, the proposed algorithm achieves a $3.2\%$ improvement in final accuracy, demonstrating the effectiveness of the UAV-EV association algorithm. Moreover, Strategy 1 shows a $4.5\%$ accuracy improvement compared to Strategy 3, further confirming the effectiveness of sharing feature extractors among related tasks. Strategy 2 outperforms Strategies 1 and 3 in early training accuracy by accelerating model convergence through frequent interactions between each EV and its nearest UAVs. However, this approach weakens generalization, resulting in lower final accuracy. Figure $2(b)$ resents the average task performance variance across all rounds for the four strategies, demonstrating that the proposed algorithm achieves a better performance balance compared to the baseline strategies.
\begin{figure}[htb]
	\centering
	\begin{minipage}{0.49\linewidth}
		\centering
		\includegraphics[width=\textwidth]{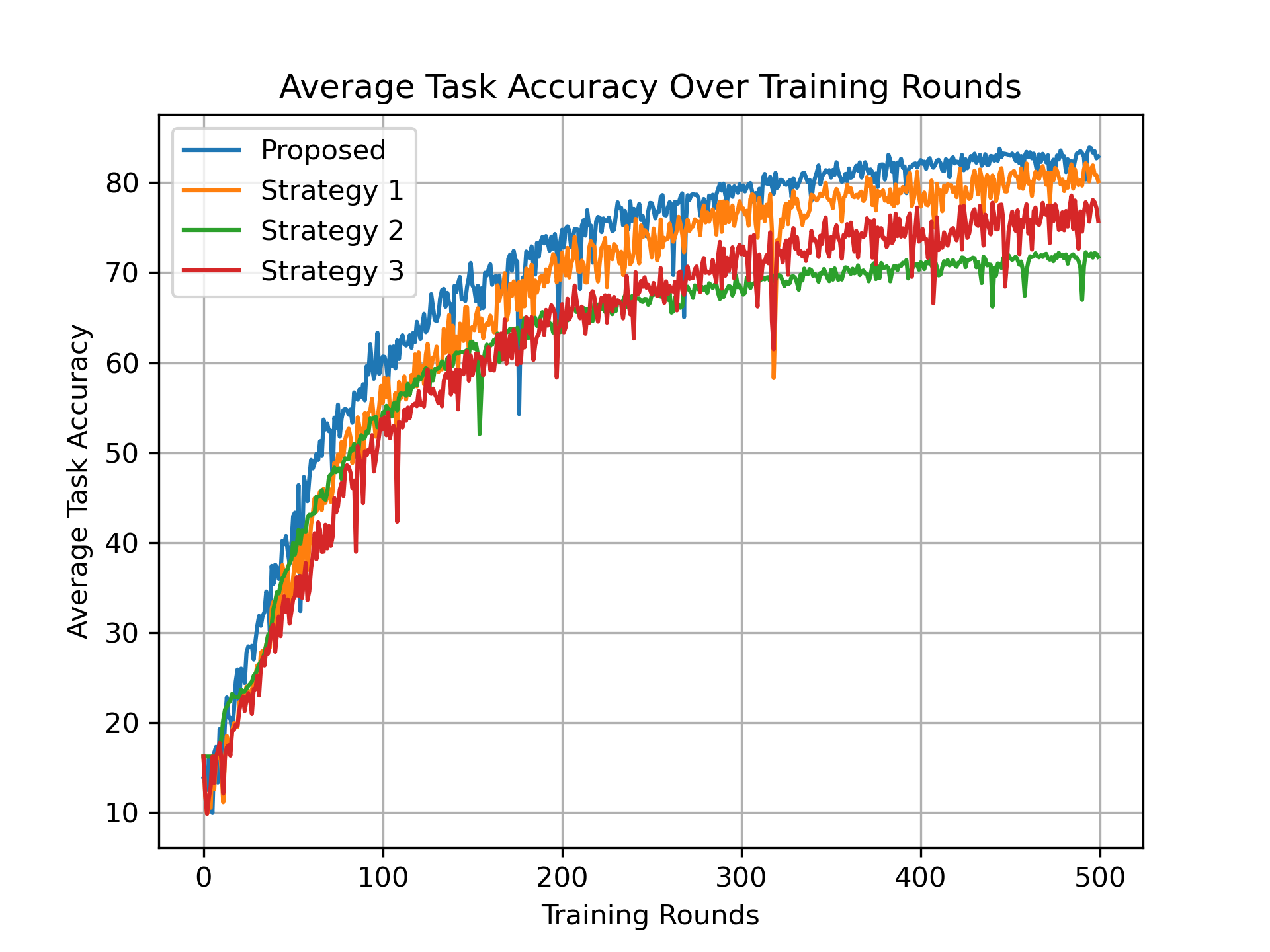}
		\subcaption{Average Accuracy}
	\end{minipage}
	\begin{minipage}{0.49\linewidth}
		\centering
		\includegraphics[width=\textwidth]{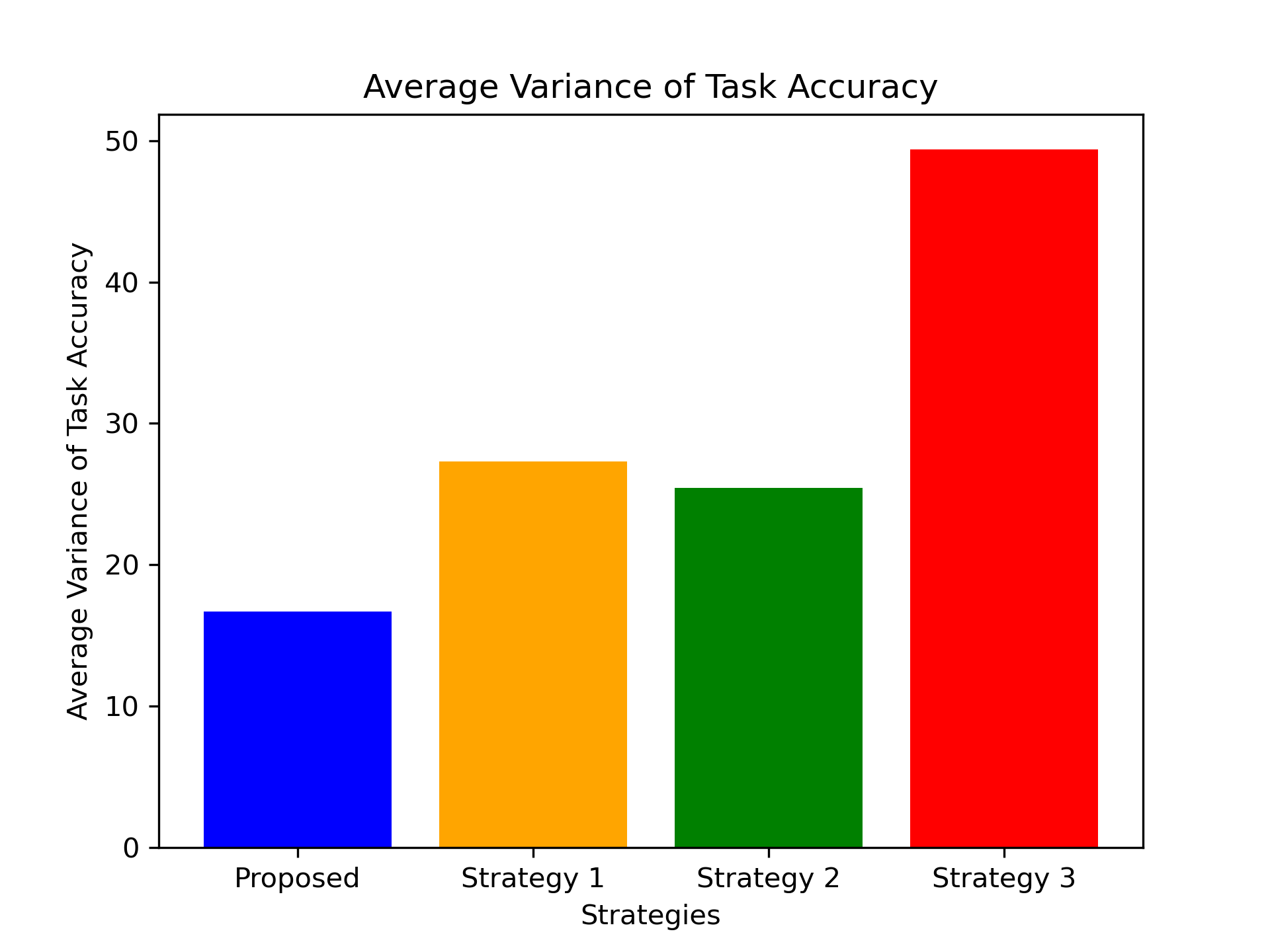}
		\subcaption{Variance}
	\end{minipage}
 
	\caption{Average Task Accuracy and Variance on MNIST}
	\label{fig:avg_acc_variance}
\end{figure}
Additionally, to investigate the impact of varying degrees of non-iid UAV data, we compare the proposed scheme with Strategy 1 to demonstrate the effectiveness of the UAV-EV association algorithm, and with Strategy 3 to highlight the benefits of sharing feature extractors. Figure $3(a)$ shows the percentage performance improvement of the proposed scheme relative to the other two strategies as $\alpha_2$ varies, with $\alpha_1 = 1$ fixed. The results show that improvement decreases as $\alpha_2$ increases, indicating the proposed scheme performs better with higher levels of non-iid data. This confirms that sharing feature extractors enhances the model’s robustness and generalization in non-iid scenarios.


\textbf{Training time comparison} To verify the effectiveness of the proposed algorithm in reducing total time consumption, we set $\lambda=0.5$ and compare the proposed algorithm with four benchmark strategies. The four benchmark strategies adopt random or nearest association, average or optimal bandwidth allocation respectively. The results in Figure $3(b)$ show that, except for the Distance-Opt strategy with the goal of time minimization, the proposed algorithm is significantly better than the other strategies under different numbers of UAVs, proving the effectiveness and optimality of the proposed joint bandwidth allocation and UAV-EV association algorithm.
\begin{figure}[htb]
	\centering
	\begin{minipage}{0.49\linewidth}
		\centering
		\includegraphics[width=\textwidth]{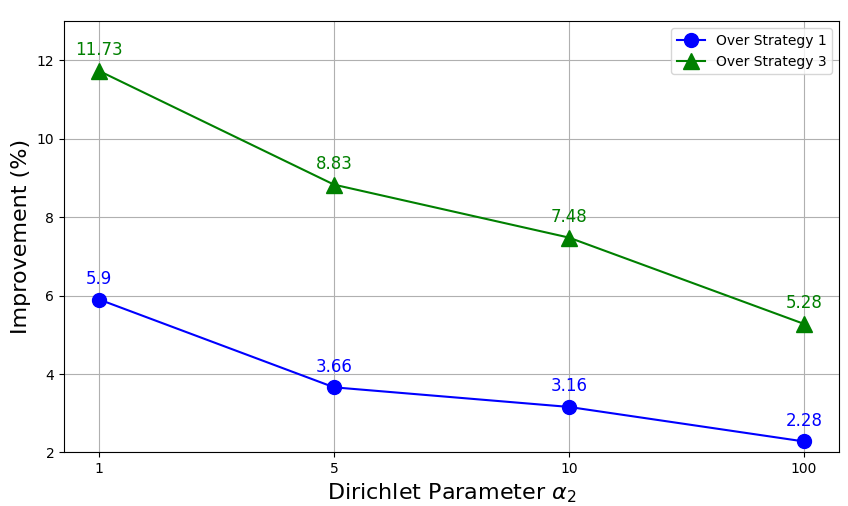}
		\subcaption{Accuracy Improvement}
	\end{minipage}
	\begin{minipage}{0.49\linewidth}
		\centering
		\includegraphics[width=\textwidth]{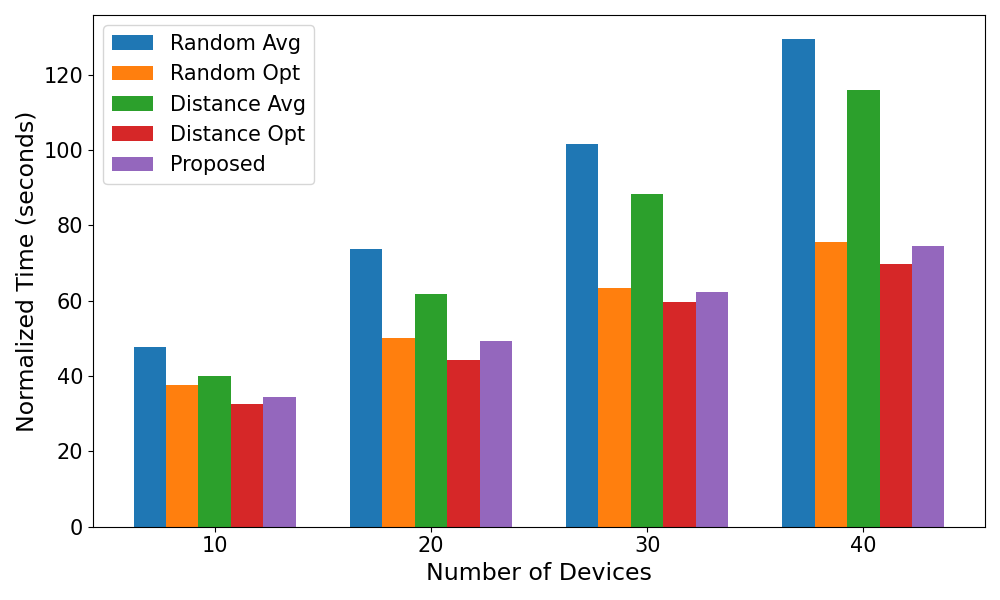}
		\subcaption{Total Time}
	\end{minipage}

	\caption{Accuracy Improvement and Total Time Consumption Comparison}
	\label{fig:example}
\end{figure}

\section{Conclusion and Future Work}
In this paper, we introduced a multi-task federated learning scheme based on task knowledge sharing. By sharing feature extractors among multiple related tasks, the overall task training acceleration and performance improvement were achieved. In addition, by analyzing the convergence of the algorithm, the joint UAV bandwidth allocation and UAV-EV association algorithm were designed to enhance training performance and reduce training time for each round. The simulation results confirm the effectiveness of the proposed algorithm. In the future, we plan to explore the underlying mechanisms for improving task performance through shared feature extractors among related tasks. Additionally, we will investigate effective methods for quantifying task correlations and design strategies to encourage collaboration in scenarios with dynamic task relevance.

\vspace{12pt}

\end{document}